\documentclass{ifacconf}
\usepackage{graphicx}      
\usepackage{natbib}        
\usepackage{amsmath}
\usepackage{amssymb}
\usepackage{svg}
\usepackage{adjustbox}

\graphicspath{{graphics/}}
\usepackage{siunitx}
\usepackage{bm}
\usepackage{algorithm}
\usepackage{algpseudocode}
\usepackage{url}
\begin{document}
\begin{frontmatter}

\title{GPU-Accelerated Motion Planning of an Underactuated Forestry Crane in Cluttered Environments}


\author[First,Second]{Minh Nhat Vu}
\author[First]{Gerald Ebmer} 
\author[First]{Alexander Watcher}
\author[First,Second]{Marc-Philip Ecker}
\author[Third]{Giang Nguyen}
\author[Second]{Tobias Glueck}

\address[First]{Automation \& Control Institute (ACIN), TU Wien, Vienna, Austria}
\address[Second]{Center for Vision, Automation \& Control,
AIT Austrian Institute of Technology GmbH, Vienna, Austria}
\address[Third]{Institute for Artificial Intelligence, University of Bremen, Germany}
\begin{abstract}                
Autonomous large-scale machine operations require fast, efficient, and collision-free motion planning while addressing unique challenges such as hydraulic actuation limits and underactuated joint dynamics. This paper presents a novel two-step motion planning framework designed for an underactuated forestry crane. The first step employs GPU-accelerated stochastic optimization to rapidly compute a globally shortest collision-free path. The second step refines this path into a dynamically feasible trajectory using a trajectory optimizer that ensures compliance with system dynamics and actuation constraints. The proposed approach is benchmarked against conventional techniques, including RRT-based methods and purely optimization-based approaches. Simulation results demonstrate substantial improvements in computation speed and motion feasibility, making this method highly suitable for complex crane systems.
\end{abstract}

\end{frontmatter}

\section{Introduction}
Motion planning for underactuated manipulators, such as forestry cranes, is inherently challenging due to the complexities of underactuated degrees of freedom, hydraulic actuation limitations, and the need to operate effectively within unstructured environments. These factors are particularly significant in the context of timber and forestry operations, where skilled workers often require the manual operation of cranes. 
However, there is a growing shortage of skilled labor, see \cite{lideskog2020opportunities}, leading to decreased productivity and operational efficiency.
Automation in the forestry industry is still underdeveloped compared to other sectors. The development of autonomous forestry cranes capable of real-time motion planning and execution, e.g., \cite{la2023framework} and \cite{la2021study}, is crucial to address these issues. Such systems must be able to operate in complex environments, avoid collisions, and ensure that operations adhere to dynamic and kinematic constraints. Conventional methods can struggle to provide solutions that are effective for real-time operation and capable of accommodating the complexity posed by underactuated systems.

\begin{figure}[htb!]
    \centering
    \def\svgwidth{1\columnwidth}
    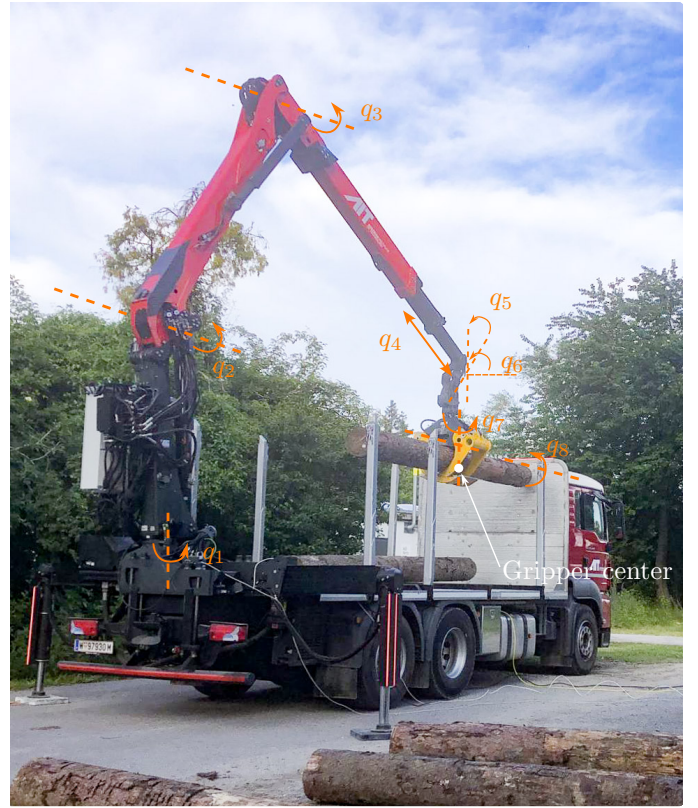
    \caption{Forestry crane considered in this work.}
    \label{fig: kinematics chain}
\end{figure}

\subsection{Related Work}
Motion planning is a fundamental task in robotics that involves generating feasible paths or trajectories to connect a given start configuration to a desired goal while satisfying system constraints. Motion planning techniques are broadly categorized into sampling-based and optimization-based approaches.

Sampling-based methods, such as Probabilistic Roadmaps (PRMs) \cite{kavraki1996probabilistic} and Rapidly-Exploring Random Trees (RRTs) \cite{luders2010bounds}, explore the configuration space by randomly sampling points and connecting them to construct feasible paths. More advanced variants, including Informed RRT* \cite{gammell2014informed} and Batch Informed Trees \cite{gammell2020batch}, improve cost efficiency while achieving asymptotic optimality. Although widely used in various applications \cite{hartmann2022effort, thomason2022task,vu2022sampling}, these methods often overlook dynamic boundary conditions, which are typically nonlinear. Consequently, additional post-processing steps, such as trajectory smoothing using spline functions, are required to ensure compatibility with higher-order derivatives.

In contrast, optimization-based motion planning techniques, such as Covariant Hamiltonian Optimization for Motion Planning (CHOMP) \cite{ratliff:2009} and TrajOpt \cite{schulman:2014}, directly optimize a discretized trajectory while enforcing system constraints and bounds. However, these methods are highly sensitive to the initial trajectory guess, making them less effective in cluttered environments with obstacles unless a precomputed global path is provided as initialization \cite{ichnowski2020deep}; \cite{vu2020fast}, \cite{vu2021fast}.

Optimization-based methods in large-scale robots are often applied for a simplified robot model, see, e.g., \cite{meiringer2019time}. 
Model Predictive Control (MPC) is a widely used technique for tracking predefined trajectories, see \cite{jebellat:2023}. These approaches often assume the availability of a global, collision-free path, limiting their applicability in complex, unstructured environments. RRT*-based planners are used for large-scale robots, see \cite{ayoub:2024}. Here, RRT* generates joint trajectories that avoid collisions but typically neglect the dynamic constraints of the system and state limits. Recently, Dynamic Motion Primitives (DMP) have been employed as motion planning to automate the forestry crane, see \cite{la2021study}, but obstacle avoidance was not addressed in this approach.
In our recent work \cite{ecker:2025}, we introduced a two-step, CPU-based trajectory optimization framework. In contrast, this study presents a flexible, GPU-based approach that efficiently supports batch generation of trajectories. While the framework in \cite{ecker:2025} remains the preferred choice for producing a single high-quality trajectory, the GPU-based method proposed here offers greater flexibility and speed when generating multiple trajectories in parallel.
\subsection{Contribution}
Planning efficient and collision-free motion for underactuated systems such as forestry cranes is particularly challenging due to their passive dynamics, hydraulic actuation limits, and the need to navigate complex environments. Many existing approaches struggle to provide computational efficiency and dynamically feasible solutions, especially in real-time applications.

This paper introduces an innovative two-step motion planning strategy for autonomous forestry cranes. The first step rapidly computes a globally shortest, collision-free path using GPU-accelerated stochastic optimization inspired by the Covariance Matrix Adaptation Evolution Strategy (CMA-ES). The second step transforms this path into a dynamically feasible trajectory with a fast optimizer that satisfies all kinematic and dynamic constraints.
The proposed approach addresses a critical gap in the motion planning domain by combining global path optimization with rapid trajectory refinement. 
Our contributions are listed as follows:
\begin{itemize}
    \item Fast GPU-based collision checking for the inverse kinematics module, global path optimization module, and optimization-based trajectory planning.
    \item Extensive simulation results show a case of the effectiveness of the proposed framework, which can consistently compute the optimal trajectories within \SI{2}{\second}. 
\end{itemize}

\section{Mathematical Modeling}\label{sec:Model}
Figure~\ref{fig: kinematics chain} depicts the forestry crane considered in this work. The configuration spaces consist of $n_d = 8$ degrees-of-freedom $\mathbf{q}^\mathrm{T}  = [\mathbf{q}_{a,b}^\mathrm{T},\mathbf{q}_p^\mathrm{T}, \mathbf{q}_{a,g}^\mathrm{T}]\in \mathbb{R}^8$, with $\mathbf{q}_{a,b}^\mathrm{T} = [{q}_1,...,{q}_4]$ (boom), $\mathbf{q}_{a,g}^\mathrm{T} = [{q}_7,q_8]$ (gripper), and $\mathbf{q}_p^\mathrm{T} = [q_5,q_6]$ (passive joint). Note that the subscripts $_a$ and $_p$ denote the actuated and passive joints, respectively. All joints are rotary, except for prismatic joint $q_4$. Using the Denavit-Hartenberg (DH) transformation, the forward kinematics $\mathrm{FK}(\mathbf{q})$ of the gripper center point can be expressed as the homogeneous transformation matrix
\begin{equation}
\mathbf{H}_g = \begin{bmatrix}
    \mathbf{R}_{0}^{g} & \mathbf{p}_{0}^{g} \\
    \mathbf{0} & 1 
\end{bmatrix}    = \mathrm{FK}(\mathbf{q}) \ .
\end{equation}
A detailed formulation of the forward kinematics for the considered crane is provided in \cite{ecker2022iterative}.



\subsection{Dynamic Model}

The system dynamics of the considered forestry crane are expressed as
\begin{equation}
    \mathbf{M}(\mathbf{q}) 
    \ddot{\mathbf{q}}
    + \mathbf{b}(\mathbf{q},\dot{\mathbf{q}}) = 
        \bm{\tau}    \ ,
\end{equation}
where the mass matrix $\mathbf{M}$, the nonlinear term $\mathbf{b}$, and the $\bm{\tau}$ vector are partitioned into 
\begin{equation}
    \begin{bmatrix}
        \mathbf{M}_{a,b} &  \mathbf{M}_{a,b,p} &  \mathbf{M}_{a,g,p}\\
        \mathbf{M}_{a,b,p} & \mathbf{M}_{p} & \mathbf{M}_{p,g,b} \\
        \mathbf{M}_{a,g,p} & \mathbf{M}_{p,g,b} & \mathbf{M}_{a,g} \\
    \end{bmatrix}, \:
        \mathbf{b} = \begin{bmatrix}
            \mathbf{b}_{a,b} \\
            \mathbf{b}_{p} \\
            \mathbf{b}_{a,g}
        \end{bmatrix}
        ,\:
        \bm{\tau} = \begin{bmatrix}
            \bm{\tau}_{a,b} \\
            \mathbf{0} \\
            \bm{\tau}_{a,g}   
        \end{bmatrix}
\end{equation}
for the individual parts $\mathbf{q}_{a,b}$, $\mathbf{q}_{a,b}$, and $\mathbf{q}_{p}$, respectively. 


In a real forestry crane, subordinate velocity controllers are employed. 
Thus, we utilize the accelerations \(\mathbf{u}^\mathrm{T} = [\ddot{\mathbf{q}}_{a,b}^\mathrm{T}, \ddot{\mathbf{q}}_{a,g}^\mathrm{T}]\) of the actuated joints as control inputs for the motion planning process. 
This formulation simplifies the forward dynamics to 
\begin{equation}
\dot{\mathbf{x}} = \mathbf{f}(\mathbf{x},\mathbf{u}) \ ,
\label{eq: state space 2}
\end{equation}
with state vector $\mathbf{x}^\mathrm{T} =
\begin{bmatrix}
\mathbf{q}^\mathrm{T} , \dot{\mathbf{q}}^\mathrm{T} 
\end{bmatrix}
$ and right-hand side 

\begin{equation}
\mathbf{f}(\mathbf{x},\mathbf{u}) = 
\begin{bmatrix}
\dot{\mathbf{q}} \\
\ddot{\mathbf{q}}_{a,b} \\
-\mathbf{M}_{u}^{-1}\left(
    \mathbf{M}_{a,b,p} \ddot{\mathbf{q}}_{a,b} + \mathbf{M}_{a,g,p} \ddot{\mathbf{q}}_{a,g}
    + \mathbf{b}_p
    \right) \\
    \ddot{\mathbf{q}}_{a,g}
\end{bmatrix} \ .
\label{eq: state space 2before}
\end{equation}
\section{Method}
The motion planning task is outlined as follows. When receiving the desired grasp pose
\begin{equation}
    \mathbf{H}_d =  
    \begin{bmatrix}
  \mathbf{R}_d & \mathbf{p}_d \\
    \mathbf{0} & 1
    \end{bmatrix} \ ,
\end{equation}
we employ the collision-free inverse kinematics method described in Subsection \ref{sec: collision-free IK} to compute the desired joint configuration $\mathbf{q}_d$, ensuring that $\mathrm{FK}(\mathbf{q}_d) = \mathbf{H}_d$ holds, while avoiding obstacles. Subsequently, the two-step trajectory optimization approach presented in Subsection ~\ref{sec: two steps} is applied to generate the optimal trajectory.

\subsection{GPU-Based Collision Checking}
We leverage GPU-accelerated collision checking to enable fast and efficient collision detection, utilizing the parallel processing capabilities of SIMD (Single Instruction, Multiple Data) hardware, cf. \cite{tang2011fast}. This approach efficiently handles large batches of parallel physical scenes, allowing us to perform numerous collision checks simultaneously. By offloading collision detection tasks to the GPU, we can significantly improve performance, especially when simulating complex environments with multiple objects, where the number of collision pairs can grow rapidly. 

For fast and efficient simulation, we utilize MuJoCo's mjx implementation from \cite{todorov2012mujoco}, which incorporates the Separating Axis Theorem (SAT) for collision detection, see \citep{gottschalk1996separating}. SAT is a well-established algorithm used to detect intersections between convex shapes. The core idea of SAT is based on the principle that if two convex shapes are disjoint, there exists at least one axis along which their projections will not overlap. Conversely, if any projection along a given axis shows an overlap, the objects are considered to collide. This is an efficient way to determine whether two objects intersect because testing projections onto various axes can quickly identify disjoint objects, see \citep{liang2015research}.

Axes candidates to project onto are the normals of the edges of the convex mesh polygons. If projections along these axes do not overlap, the objects are disjoint. The number of candidates is proportional to the number of edges in the polygons, which means SAT can be computationally expensive when applied to meshes with many edges or vertices. However, SAT is particularly well-suited for GPU implementation because the calculations involved, projecting vertices onto axes and testing for overlap, are independent for each axis and can be parallelized efficiently. This drastically improves computational speed compared to a CPU-based approach, cf. \citep{liang2015research}.

Using GPU-based collision check, we obtain a signed distance $d(\mathbf{q})$ between the robot's surface and the obstacles in which $d<0$ indicates the robot is in a collision. 

\subsection{Collision-Free Inverse Kinematics}
\label{sec: collision-free IK}

To align the end-effector pose $\mathbf{H}_g$ with the desired pose $\mathbf{H}_d$, we minimize residual factors combining positional, rotational, regularization, and collision-avoidance terms. The optimization problem is formulated as
\begin{equation}\label{eq: IK opt}
\mathbf{q}_d = \arg\min_{\mathbf{q}} ||\mathbf{r}||_2^2 \ ,
\end{equation}
with $ \mathbf{r}^\mathrm{T} = \begin{bmatrix}
        \mathbf{r}_p^\mathrm{T}  &
        {r}_q &
        {r}_{{c}} &
        {r}_{e}
    \end{bmatrix} \ .
$
The position residual, $\mathbf{r}_p = \mathbf{p}_g - \mathbf{p}_d$, penalizes deviations in the position of the end-effector. The orientation residual is computed as
\begin{equation}
{r}_q = \lambda_q \cdot \mathrm{Log}(\mathbf{R}_d^\mathrm{T} \mathbf{R}_g) \ ,
\end{equation}
where $\mathrm{Log}(\cdot)$ denotes the logarithmic map on $SO(3)$, and $\lambda_q$ is a scaling factor to balance orientation and position terms. 
Collision avoidance is incorporated through the residual ${r}_{{c}}$, which penalizes configurations that lead to self-collision or collision with the environment. 
This term is defined as
\begin{equation}
{r}_{{c}} = \lambda_c \cdot  \text{ReLU}(- d(\mathbf{q}))\ ,
\end{equation}
where $d(q)$ is the signed distance and $\lambda_c$ is a scaling factor. Lastly, ${r}_e$ is the term ensuring the equilibrium of the two unactuated angles, computed as 
\begin{equation}
{r}_e = \lambda_e \cdot  ||\mathbf{g}_p||_2^2 \ ,
\end{equation}
where $\mathbf{g}_p \in \mathbb{R}^2$ is the gravitational vector of passive joint $\mathbf{q}_p$ and $\lambda_e$ is a scaling factor. 

\section{Spline-Based Trajectory Optimization Framework}
\label{sec: two steps}

\subsection{Stochastic-Based Path Optimization}
Once the desired configuration $\mathbf{q}_d$ is determined according to \eqref{eq: IK opt}, we generate a smooth, collision-free joint-space path $\mathbf{q}(s)$, with monotonic path parameter $s\in[0,1]$, for the robotic manipulator, ensuring a continuous transition from the initial configuration $\mathbf{q}_{0}$ to the desired configuration $\mathbf{q}_d$. We parameterize the path $\mathbf{q}(s)$ as a cubic B-spline interpolation characterized by control points $\mathbf{P}_i,\: i\in n_c$, expressed in the form
\begin{equation}
\mathbf{q}(s) = \sum_{i=0}^{n_c - p - 1} \mathbf{P}_i \, N_i^p(s) \ ,
\end{equation}
where \(\mathbf{P}_i \in \mathbb{R}^{n_p}\) are the control points in joint space, \(N_i^p(s)\) are the cubic B-spline basis functions of degree \(p=3\), and \(n_c\) is the number of control points. 

An initial guess for the control points \(\{\mathbf{P}_i\}\) is often obtained by linear interpolation between \(\mathbf{q}_{0}\) and \(\mathbf{q}_{d}\). This ensures that \(\mathbf{P}_j = \mathbf{q}_{0}, \ j = 0,\dots,p \) and \(\mathbf{P}_{k} = \mathbf{q}_{{d}}, \ k = n_c-p-1, \dots, n_c -1\), while the intermediate control points are evenly spaced and subject to adjustment during optimization. For further details on B-splines, see \cite{PiR08}.

Here, the goal is to find the optimal set of control points \(\mathcal{P}^* = \{\mathbf{P}_{p+1}^*, \dots, \mathbf{P}_{n_c-p-2}^*\}\) involving the following optimization problem
\begin{subequations}
\label{eq: path opt}
\begin{alignat}{3}
    &\mathcal{P}^* = \arg\min_{\mathcal{P}} && \: ( J_{\text{p}}(\mathcal{P}) 
+ w_{\text{c}} J_{\text{c}}(\mathcal{P}) 
+ w_{\text{l}} J_{\text{l}}(\mathcal{P}))  \\
\label{eq: cost path length}
&J_{\text{p}}(\mathcal{P}) 
= \sum_{k=1}^{n_s}
&& \|\mathbf{q}(s_k) - \mathbf{q}(s_{k-1})\|^2_2\\
\label{eq: cost collision}
&J_{\text{c}}(\mathcal{P}) 
= \sum_{k=1}^{n_s} && r_c(\mathbf{q}(s_k)) \\
\label{eq: cost feasibility}
&J_{\text{l}}(\mathcal{P}) 
= \sum_{k=1}^{n_s} && \max\bigl(\mathbf{0},\, \mathbf{q}(s_k) - \mathbf{q}_{\max}\bigr)^2  \\ 
& && + \max\bigl(\mathbf{0},\,\mathbf{q}_{\min} - \mathbf{q}(s_k)\bigr)^2 \ ,
\end{alignat}
\end{subequations}
with $n_s$ is the number of discretizing point along the path, $s_k = \frac{k}{n_s}$, $\mathbf{q}_\mathrm{max}$ and $\mathbf{q}_\mathrm{min}$ are the limits of the joint angles. Note that $n_s$ is the number of checkpoints along the trajectory. 
The terms in \eqref{eq: cost path length}-\eqref{eq: cost feasibility} correspond to the path length cost, collision cost, and feasibility cost, respectively. The scaling factors $w_{\text{c}}$ and $w_{\text{l}}$ are introduced to appropriately weight each cost term.

A stochastic optimization scheme is employed to determine the B-spline control points $\mathcal{P}$. 
Algorithm~\ref{alg:cmaes} outlines the main steps inspired by the Covariance Matrix Adaptation Evolution Strategy (CMA-ES), see \cite{liang2019covariance}. 
Each iteration refines the distribution over control points until a stopping criterion is met (e.g., reaching a minimal cost or hitting a maximum iteration). The final mean $\boldsymbol{\mu}$ is reshaped into $\mathcal{P}$, which defines a smooth, collision-free trajectory via the cubic B-spline parameterization. 

\begin{algorithm}
\caption{Stochastic Optimization with Log-Rank Weights for B-Spline Control Points.}
\label{alg:cmaes}
\begin{algorithmic}[1]
\Require 
  \(\bm{\mu} \in \mathbb{R}^{\tilde{n}_c \times n_d}\) (initial mean), 
  \(\mathbf{C} \in \mathbb{R}^{(\tilde{n}_c  \times n_d) \times (\tilde{n}_c  \times n_d)}\) (initial covariance), $\tilde{n}_c = n_c - 2(d+1)$
  \(\lambda \in \mathbb{N}\) (population size), 
  \(\mu \in \mathbb{N}\) (elite count)
\Ensure 
  \(\mathbf{P}^*\) (optimized B-spline control points)

\vspace{2pt}
\Statex \textbf{repeat until convergence:}
\State \textbf{(1) Sampling:} 
   For \(i = 1, 2, \ldots, \lambda\):
   \begin{equation*}
     \mathbf{\xi}_i \sim \mathcal{N}\bigl(\bm{\mu} \ ,\, \sigma^2 \mathbf{C}\bigr),
     \quad \mathbf{\xi}_i \in \mathbb{R}^{\tilde{n}_c  \times n_d} \ .
   \end{equation*}

\State \textbf{(2) Evaluation:}
\begin{align}
\mathbf{q}^{(i)}(s)
&=\;\mathrm{B\_Spline}\Bigl(\{\mathbf{P}_0^{(i)}\ , \dots \ , \mathbf{P}_{n_c-1}^{(i)}\},\,s\Bigr) \ , 
\\[6pt]
J\bigl(\mathbf{\xi}_i\bigr)
&=\;\mathrm{Cost}\!\Bigl(\mathbf{q}^{(i)}(s)\Bigr) \ ,
\end{align}
\noindent
where control points \(\mathbf{P}_k^{(i)}\in \mathbb{R}^{n_d}\) are taken from \(\mathbf{\xi}_i\in\mathbb{R}^{\tilde{n}_c \times n_d}\),
\(\mathrm{B\_Spline}(\cdot)\) constructs the B-spline trajectory 
\(\mathbf{q}^{(i)}(s)\),
and \(\mathrm{Cost}(\cdot)\) aggregates costs in (\ref{eq: path opt}).

\State \textbf{(3) Selection:} 
   Sort all \(\{\mathbf{\xi}_i\}\) by cost \(J(\mathbf{\xi}_i)\) in ascending order.
   Let \(\mathbf{\xi}_{(1)}, \mathbf{\xi}_{(2)}, \ldots, \mathbf{\xi}_{(\nu)}\) denote 
   the \(\nu\) lowest-cost solutions (the ``elites'').

\State \textbf{(4) Log-Rank Weights:} 
   For \(j = 1, 2, \ldots, \nu\), assign
   \begin{equation*}
     w_j \;=\; 
     \frac{\ln\bigl(\nu + 0.5\bigr) \;-\; \ln(j)}
          {\sum_{k=1}^{\nu}
            \Bigl[\ln\bigl(\nu + 0.5\bigr) \;-\; \ln(k)\Bigr]} \ .
   \end{equation*}

\State \textbf{(5) Update Mean:}
   \begin{equation*}
     \bm{\mu} 
     \;\leftarrow\; 
     \sum_{j=1}^\nu \; w_j \,\mathbf{x}_{(j)} \ .
   \end{equation*}

\State \textbf{(6) Update Covariance:}
   \begin{equation*}
     \mathbf{C} 
     \;\leftarrow\; 
     \sum_{j=1}^\nu 
       w_j\,\Bigl(\mathbf{x}_{(j)} - \bm{\mu}\Bigr)\,
       \Bigl(\mathbf{x}_{(j)} - \bm{\mu}\Bigr)^\mathrm{T}
       \;+\;\epsilon\,\mathbf{I} \ ,
   \end{equation*}
   where \(\epsilon\mathbf{I}\) provides numerical stability.
\end{algorithmic}
\end{algorithm}

\subsection{Trajectory Tracking Optimization}
After obtaining the collision-free and kinematics feasible path $\mathbf{q}^*(s)$ from the optimal set of control points $\mathcal{P}^*$. The near-time optimal trajectory optimization is processed as follows.  
The optimization scheme is formulated using the direct collocation method, see, e.g., \cite{betts2010practical}, by discretizing the trajectory into $n_s$ grid points and solving the resulting static optimization problem
\begin{subequations}\label{Eq: discrete}
\begin{align}
\label{Eq: discrete a}
 \min_{\bm{\xi}} &\: J(\bm{\xi}) = t_F + \omega_s \sum_{k=1}^{n_s} \mathbf{u}_k^\mathrm{T} \mathbf{R} \mathbf{u}_k + \omega_t \sum_{k=1}^{n_s} ||\mathbf{q}_{a,k} - \mathbf{q}_{a,k}^*||^2_2\\
\label{Eq: discrete b}
\text{s.t.} \:\: &\mathbf{x}_{k+1} - \mathbf{x}_k = \dfrac{1}{2}h(\mathbf{f}_k + \mathbf{f}_{k+1}), \: k=1,\dots,N\\
\label{Eq: discrete c}
& \mathbf{x}_0 = \mathbf{x}_{S}, \:\: \mathbf{x}_N= \mathbf{x}_{T} \\
\label{Eq: discrete d}
& \underline{\mathbf{x}} \leq \mathbf{x}_k  \leq \overline{\mathbf{x}} \ , 
\end{align}
\end{subequations}
where the index $k$ indicates the discrete time step $t=k h$, with sampling time $h = t_F / n_s$, $\mathbf{f}_k = \mathbf{f}(\mathbf{x}_k,\mathbf{u}_k)$ is the right-hand side of (\ref{eq: state space 2}), and $\mathbf{q}_a^\mathrm{T} = [\mathbf{q}_{a,b}^\mathrm{T},\mathbf{q}_{a,g}^\mathrm{T}]$. 
The vector of optimization variables is given by
\begin{equation}
\bm{\xi}^\mathrm{T} = [t_F,\mathbf{x}_0^\mathrm{T}, \dots, \mathbf{x}_N^\mathrm{T},\mathbf{u}_0^\mathrm{T}, \dots, \mathbf{u}_N^\mathrm{T}] \: . 
\end{equation} 
Note that the final time $t_F$ in (\ref{Eq: discrete a}) denotes the time it takes for the system (\ref{eq: state space 2}) to transition from the starting state $\mathbf{x}_S^\mathrm{T} = [\mathbf{q}_0^\mathrm{T},\mathbf{0}^\mathrm{T}]$ to the target state $\mathbf{x}_T^\mathrm{T} = [\mathbf{q}_d^\mathrm{T},\mathbf{0}^\mathrm{T}]$. Herein, $\mathbf{q}_{a,k}^*$ is taken from the optimal path $\mathbf{q}^*(s)$. The state constraints in (\ref{Eq: discrete d}) enforce system limits, such as joint limits and hydraulic actuator limits. Furthermore, $\mathbf{R}$ is a positive definite weighting matrix applied to the control input $\mathbf{u}$.
\begin{figure}
    \centering
    \includegraphics[width=0.4\textwidth]{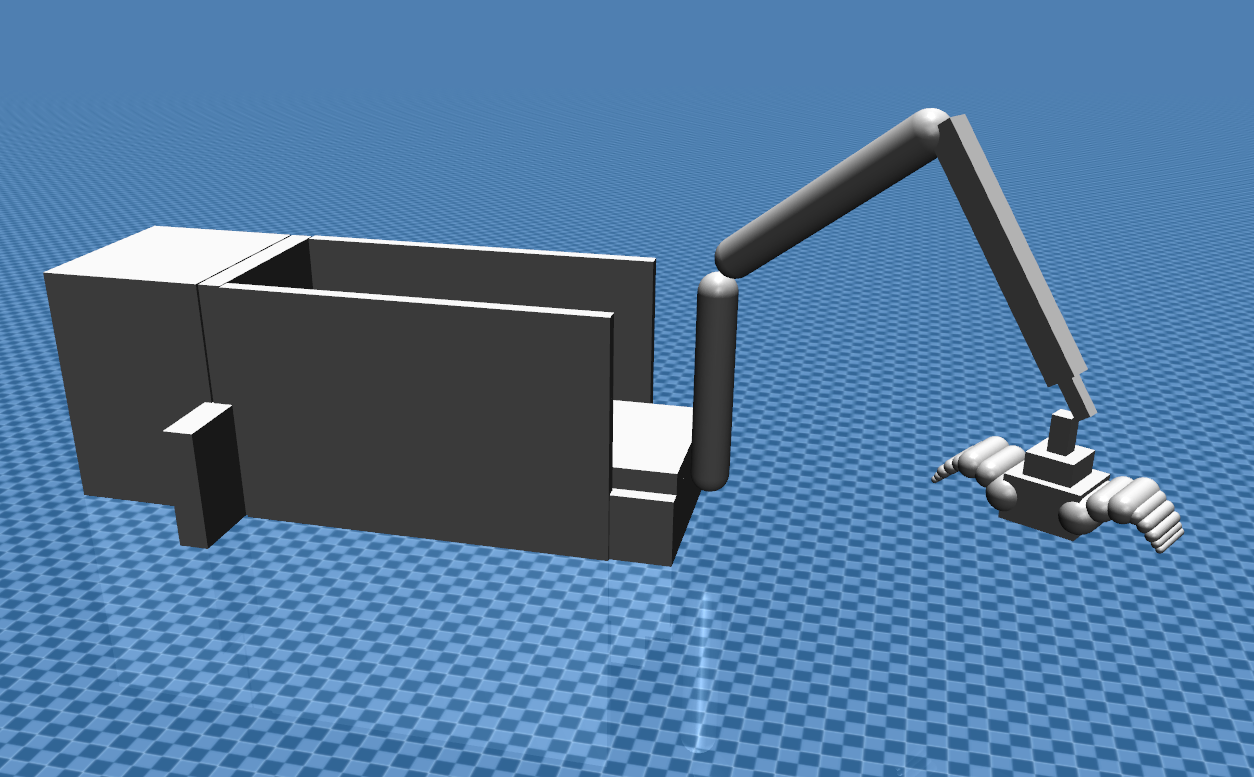} 
    \caption{Illustration of the collision meshes of the crane model.}
    \label{fig:mujoco_col} 
\end{figure}
\section{Simulation Results}
This section presents the numerical results of the proposed method and benchmarks its performance against two alternative approaches: the LQR-RRT* implementation from \cite{PythonRobotics} and the trajectory optimization formulation in (\ref{Eq: discrete}) without a precomputed global initial path. All experiments are conducted on an AMD Ryzen 9 7900 CPU @ 3.7 GHz × 12. For GPU-accelerated distance computations, we utilize MuJoCo XLA from \cite{mujoco2025}. The nonlinear optimization problem in (\ref{Eq: discrete}) is solved using the advanced interior-point solver IPOPT with the MA57 linear solver, as detailed in \cite{wachter2006implementation}. To enhance computational efficiency, we employ automatic differentiation (AD) for both the cost and constraint functions using the CasADi package \cite{andersson2019casadi}. Each trajectory is discretized into $n_s = 100$ collocation points, resulting in 16,001 optimization variables in (\ref{Eq: discrete}). Given $n_s = 100$ and a population size of 100 in Algorithm \ref{alg:cmaes}, a total of 10,000 collision checks are performed during each update of the stochastic optimization process in Algorithm \ref{alg:cmaes}.

\subsection{GPU-based Collision Checking}
While SAT is an effective method for collision detection, its computational complexity increases with mesh complexity. For intricate geometries, SAT requires evaluating a large number of potential separating axes, which can become a computational bottleneck. To address this, we simplify the meshes, reducing the number of collision pairs and potential separating axes while preserving geometric fidelity. The final model comprises 32 collision meshes, as illustrated in Figure \ref{fig:mujoco_col}, enabling efficient and robust collision detection. This optimization allows for the evaluation of 100 trajectories, each discretized into 100 points, achieving an average runtime of 0.014 seconds on an NVIDIA GeForce RTX™ 4090 for a total of 10,000 evaluations.

\subsection{Monte Carlo Simulations}
\begin{figure}
    \centering
    \def\svgwidth{1\columnwidth}
\begingroup%
  \makeatletter%
  \providecommand\color[2][]{%
    \errmessage{(Inkscape) Color is used for the text in Inkscape, but the package 'color.sty' is not loaded}%
    \renewcommand\color[2][]{}%
  }%
  \providecommand\transparent[1]{%
    \errmessage{(Inkscape) Transparency is used (non-zero) for the text in Inkscape, but the package 'transparent.sty' is not loaded}%
    \renewcommand\transparent[1]{}%
  }%
  \providecommand\rotatebox[2]{#2}%
  \newcommand*\fsize{\dimexpr\f@size pt\relax}%
  \newcommand*\lineheight[1]{\fontsize{\fsize}{#1\fsize}\selectfont}%
  \ifx\svgwidth\undefined%
    \setlength{\unitlength}{418.7812673bp}%
    \ifx\svgscale\undefined%
      \relax%
    \else%
      \setlength{\unitlength}{\unitlength * \real{\svgscale}}%
    \fi%
  \else%
    \setlength{\unitlength}{\svgwidth}%
  \fi%
  \global\let\svgwidth\undefined%
  \global\let\svgscale\undefined%
  \makeatother%
  \begin{picture}(1,0.69330507)%
    \lineheight{1}%
    \setlength\tabcolsep{0pt}%
    \put(0,0){\includegraphics[width=\unitlength,page=1]{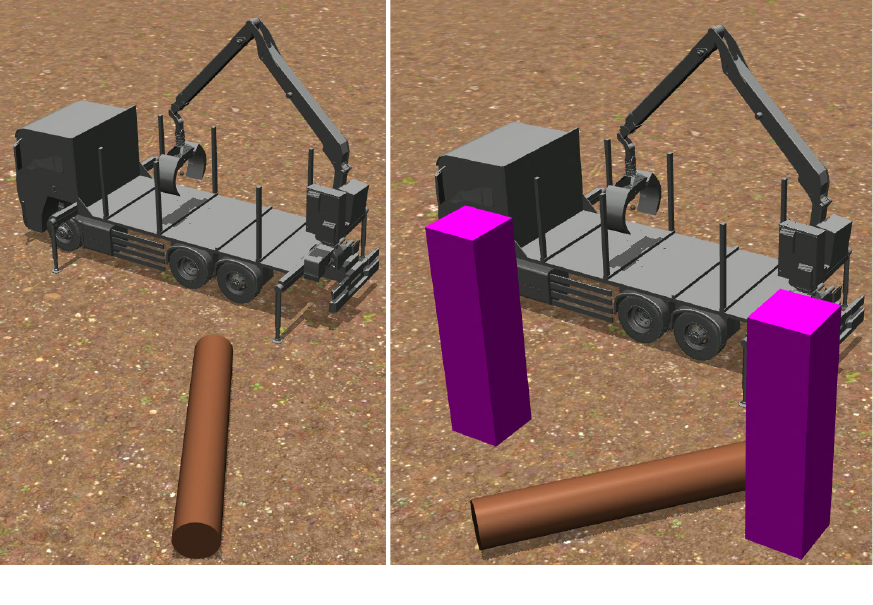}}%
    \put(0.18437357,0.0062962){\color[rgb]{0,0,0}\makebox(0,0)[lt]{\lineheight{1.25}\smash{\begin{tabular}[t]{l}(a)\end{tabular}}}}%
    \put(0.66374434,0.0064126){\color[rgb]{0,0,0}\makebox(0,0)[lt]{\lineheight{1.25}\smash{\begin{tabular}[t]{l}(b)\end{tabular}}}}%
  \end{picture}%
\endgroup%
        
    \caption{Benchmark environments. (a) Collision with the truck's trunk is considered. (b) Cluttered environment with obstacles.}
    \label{fig:Environments}
\end{figure}

\begin{figure}
    \centering
    \def\svgwidth{0.75\columnwidth}
\begingroup%
  \makeatletter%
  \providecommand\color[2][]{%
    \errmessage{(Inkscape) Color is used for the text in Inkscape, but the package 'color.sty' is not loaded}%
    \renewcommand\color[2][]{}%
  }%
  \providecommand\transparent[1]{%
    \errmessage{(Inkscape) Transparency is used (non-zero) for the text in Inkscape, but the package 'transparent.sty' is not loaded}%
    \renewcommand\transparent[1]{}%
  }%
  \providecommand\rotatebox[2]{#2}%
  \newcommand*\fsize{\dimexpr\f@size pt\relax}%
  \newcommand*\lineheight[1]{\fontsize{\fsize}{#1\fsize}\selectfont}%
  \ifx\svgwidth\undefined%
    \setlength{\unitlength}{538.83220522bp}%
    \ifx\svgscale\undefined%
      \relax%
    \else%
      \setlength{\unitlength}{\unitlength * \real{\svgscale}}%
    \fi%
  \else%
    \setlength{\unitlength}{\svgwidth}%
  \fi%
  \global\let\svgwidth\undefined%
  \global\let\svgscale\undefined%
  \makeatother%
  \begin{picture}(1,1.16923071)%
    \lineheight{1}%
    \setlength\tabcolsep{0pt}%
    \put(0,0){\includegraphics[width=\unitlength,page=1]{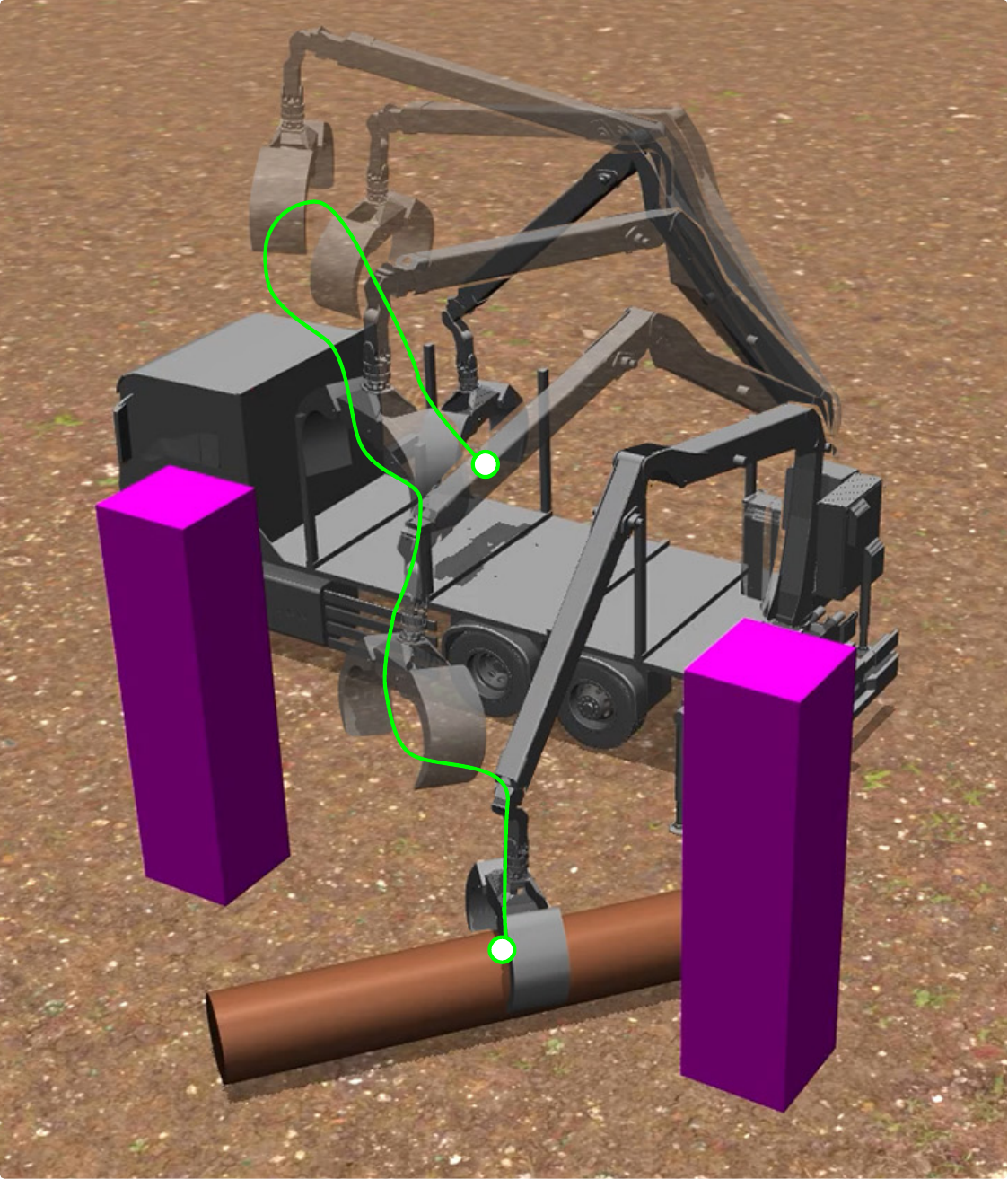}}%
    \put(0.51730773,0.67692302){\color[rgb]{0,0,0}\makebox(0,0)[lt]{\lineheight{1.25}\smash{\begin{tabular}[t]{l}\textcolor{white}{init state}\end{tabular}}}}%
    \put(0.07639509,1.02102806){\color[rgb]{0,0,0}\makebox(0,0)[lt]{\lineheight{1.25}\smash{\begin{tabular}[t]{l}\textcolor{white}{end-effector path}\end{tabular}}}}%
    \put(0.32846114,0.24419215){\color[rgb]{0,0,0}\makebox(0,0)[lt]{\lineheight{1.25}\smash{\begin{tabular}[t]{l}\textcolor{white}{target state}\end{tabular}}}}%
    \put(0,0){\includegraphics[width=\unitlength,page=2]{overlay-crane.pdf}}%
  \end{picture}%
\endgroup%
        
    \caption{Snapshots of the collision-free motion of the forestry crane in a cluttered environment.}
    \label{fig:snapshots}
\end{figure}

We evaluate our approach in two distinct environments, illustrated in Figure~\ref{fig:Environments}. The first environment, Figure~\ref{fig:Environments} (a), focuses on self collisions with rungs on the loading area. In the second scenario, Figure~\ref{fig:Environments} (b), the forestry crane has to avoid the trunk of the truck and two other obstacles. In each environment, we randomly generate the $10^3$ locations of the wood log in a collision-free space. A collision-free feasible trajectory that guides the forestry crane maneuvers in the cluttered environment is shown in Figure \ref{fig:snapshots}.  

The LQR-RRT* algorithm and the trajectory optimization without collision-free path initialization are subjected to the same time constraint of \SI{90}{\second} across all test cases. For the LQR-RRT* algorithm, a diagonal state weighting matrix, $\mathbf{Q}$, is used. The diagonal entries are set to $1$ for $\mathbf{q}_a$, $0.01$ for both $\mathbf{q}_p$ and $\dot{\mathbf{q}}_a$, and $0.01$ for $\dot{\mathbf{q}}_p$. To prevent large accelerations in the control input $\mathbf{u}$, the control weighting matrix is defined as $\mathbf{R} = 0.01\mathbf{I}$. 

The results in Table \ref{tab: comparison} show the performance of the proposed approach compared to LQR-RRT* and trajectory optimization. In this Monte Carlo simulation with $10^3$ trials, our approach requires less than $\SI{2}{\second}$ to compute a near-time optimal trajectory, while other methods require significantly more computation time. If we only use trajectory optimization without the collision-free shortest path as a guide, longer travel times can be expected. The large standard deviations in the trajectory times of LQR-RRT* indicate that the results are still far from optimal, even after 30 seconds. In contrast, the proposed approach achieves trajectories with significantly lower average durations and less variability, emphasizing its reliability and efficiency. 
Note that we do not count the inverse kinematics' computation time for the LQR-RRT* and the trajectory optimization problem. The breakdown of the average computation times for each submodule in our approach is shown in Table \ref{tab: run time}. A video of the crane's movement in the simulated environment is provided at \url{https://www.acin.tuwien.ac.at/7e94/}.

\begin{table*}[htb!]
    \centering
    \caption{Success rates and trajectory durations of the global planners in $10^3$ runs.}\label{tab:StatisticsVPSTO}
    \begin{tabular}{|c|c|c|c||c|c||c|c|c|c||}
        \hline
         \multicolumn{1}{||c|}{\textbf{Env.}}  & \multicolumn{3}{|c||}{\textbf{Success Rate in \%}} & \multicolumn{3}{|c||}{\textbf{Traverse time in s}} & \multicolumn{3}{|c||}{\textbf{Computation time in s}}\\
         \hline\hline
         &Ours  & LQR-RRT* & Traj.Opt. & Ours & LQR-RRT* &Traj.Opt&Ours  & LQR-RRT* & Traj.Opt.\\
        \hline
        1 &  100 & 45.7
                & 82.2 & 7.54$\pm$4.23
                & 25.20$\pm$5.27 & 8.29$\pm$3.15 & 1.42$\pm 0.12$  & 32.15 $\pm$ 27.34 & 11.25$\pm$ 3.75 \\
        \hline\hline
        2 &  100 & 40.1 & 72.7
                  & 7.58 $\pm$3.24 & 30.72$\pm$ 4.38
                  & 7.97$\pm$2.31 & 1.51$\pm$0.18& 42.35$\pm$12.5 & 15.72$\pm$4.17\\
        \hline
    \end{tabular}
    \label{tab: comparison}
\end{table*}

\begin{table}[H]
    \centering
    \caption{Average computation time of each submodule over $10^3$ runs.}\label{tab:StatisticsVPSTO}
    \begin{tabular}{|c|c|c|c|c}
    \hline
     & IK (\ref{eq: IK opt}) & Path Opt. (\ref{eq: path opt})& Traj. Opt. (\ref{Eq: discrete}) \\
     \hline
    Run time in \SI{}{\second} & 0.005 & 0.96 & 0.51 \\
    \hline
    \end{tabular}
    \label{tab: run time}
\end{table}

\section{Conclusion}
This paper introduces a comprehensive framework for fast computing a collision-free and dynamically feasible trajectory for a forestry crane. The approach integrates stochastic path optimization to generate an initial collision-free path, which then serves as a guide for the subsequent optimization-based planner. The proposed framework is evaluated through Monte Carlo simulations, demonstrating its effectiveness. Compared to LQR-RRT* and purely optimization-based planners, our method achieves significantly faster computation times while maintaining high solution quality, making it well-suited for complex motion planning tasks.

\bibliography{ifacconf}                                                                  
\end{document}